\documentclass[sigconf]{acmart}

\usepackage{booktabs}
\usepackage{multirow}
\usepackage{graphicx}
\usepackage{color}
\usepackage{caption}
\usepackage{subcaption}

\AtBeginDocument{%
  \providecommand\BibTeX{{%
    \normalfont B\kern-0.5em{\scshape i\kern-0.25em b}\kern-0.8em\TeX}}}

\setcopyright{acmcopyright}
\copyrightyear{2018}
\acmYear{2018}
\acmDOI{10.1145/1122445.1122456}

\acmConference[Woodstock '18]{Woodstock '18: ACM Symposium on Neural
  Gaze Detection}{June 03--05, 2018}{Woodstock, NY}
\acmBooktitle{Woodstock '18: ACM Symposium on Neural Gaze Detection,
  June 03--05, 2018, Woodstock, NY}
\acmPrice{15.00}
\acmISBN{978-1-4503-XXXX-X/18/06}

\newcommand{\quotes}[1]{``#1''}

\newcommand{\parencite}[1]{\cite{#1}}
\newcommand{\meanstd}[2]{${#1}_{\pm #2}$}

\begin{document}

\title{Catastrophic Forgetting in Deep Graph Networks: an Introductory Benchmark for Graph Classification}

\author{Antonio Carta}
\authornote{Authors contributed equally to this research.}
\affiliation{%
  \institution{Department of Computer Science - University of Pisa}
  \streetaddress{Largo B. Pontecorvo, 3}
  \city{Pisa}
  \country{Italy}
  \postcode{56127}
}
\email{antonio.carta@di.unipi.it}
\author{Andrea Cossu}
\authornotemark[1]
\affiliation{
  \institution{Scuola Normale Superiore}
  \streetaddress{Piazze dei Cavalieri, 7}
  \city{Pisa}
  \country{Italy}
  \postcode{56126}
}
\email{andrea.cossu@sns.it}
\author{Federico Errica}
\authornotemark[1]
\affiliation{%
  \institution{Department of Computer Science - University of Pisa}
  \streetaddress{Largo B. Pontecorvo, 3}
  \city{Pisa}
  \country{Italy}
  \postcode{56127}
}
\email{federico.errica@phd.unipi.it}
\author{Davide Bacciu}
\affiliation{%
  \institution{Department of Computer Science - University of Pisa}
  \streetaddress{Largo B. Pontecorvo, 3}
  \city{Pisa}
  \country{Italy}
  \postcode{56127}
}
\email{bacciu@di.unipi.it}

\renewcommand{\shortauthors}{A. Carta, A. Cossu, F. Errica, and D. Bacciu}

\begin{abstract}
In this work, we study the phenomenon of catastrophic forgetting in the graph representation learning scenario.  The primary objective of the analysis is to understand whether classical continual learning techniques for flat and sequential data have a tangible impact on performances when applied to graph data. To do so, we experiment with a structure-agnostic model and a deep graph network in a robust and controlled environment on three different datasets. The benchmark is complemented by an investigation on the effect of structure-preserving regularization techniques on catastrophic forgetting. We find that replay is the most effective strategy in so far, which also benefits the most from the use of regularization. Our findings suggest interesting future research at the intersection of the continual and graph representation learning fields. Finally, we provide researchers with a flexible software framework to reproduce our results and carry out further experiments.
\end{abstract}




\maketitle

\section{Introduction}
\label{sec:introduction}
Building a robust machine learning model that incrementally learns from different tasks without forgetting requires methodologies that account for drifts in the input distribution. The Continual Learning (CL) research field addresses the catastrophic forgetting problem \cite{grossberg1980, French1999a} by devising learning algorithms that improve a model's ability to retain previously gathered information. As of today, CL methods have been studied from the perspective of flat data \cite{Kirkpatrick2017a, Maltoni2018a, Shin2017a} and, to a lesser extent, sequential data \cite{Sodhani2020a, ehret2020a}.

Graph Representation Learning (GRL) is the study of machine learning models that can make predictions about input data represented as a graph. GRL methods naturally find application in social sciences \parencite{nechaev_sociallink_2018}, recommender systems \parencite{bobadilla_recommender_2013}, cheminformatics \parencite{micheli_introduction_2007}, security \parencite{iadarola_graph-based_2018} and natural language processing \cite{marcheggiani_exploiting_2018}, where the data is arbitrarily structured and cycles may occur \cite{micheli_introduction_2007}.

At present, the literature lacks an analysis of catastrophic forgetting in models that deal with graphs. The few existing works focus on new approaches which are not compared to existing CL strategies on challenging benchmarks \cite{wang2020, zhou2021}. This work makes the first step in this direction by carrying out continual learning experiments on graph classification benchmarks in a robust and controlled framework. In this context, we investigate whether specific GRL regularization strategies can mitigate catastrophic forgetting by enforcing structural information preservation.

Our contribution is two-fold. First of all, we study whether CL techniques for flat data still work on the graph domain. If that is not the case, the results will call for different and novel approaches to be developed. Secondly, we provide a robust and reproducible framework to carry out Continual Learning experiments on graph-structured data. Indeed the GRL field has suffered serious reproducibility issues that impacted chemical and social benchmarks \parencite{errica_fair_2020}. By publicly releasing our code, we foster this trend to prevent common malpractices such as the usage of custom data splits, the absence of a model selection, and incorrect evaluations of the estimated risk on a validation (rather than test) set.
\section{Related Works}
\label{sec:related-works}
This section introduces the high-level notions of both continual learning and deep learning for graphs. 

\subsection{Continual Learning}
\label{subsec:continual-learning}
The main objective of CL is to learn from a continuous stream of data while mitigating catastrophic forgetting of previously acquired knowledge \cite{Parisi2019a}. Continual learning models can be roughly categorized into three families: regularization strategies, architectural strategies and replay strategies. Though not entirely comprehensive, this taxonomy includes most of the currently used CL strategies. \\
\textbf{Regularization strategies} add a penalization to the standard loss function to enforce the stability of existing parameters. Elastic Weight Consolidation (EWC) \parencite{Kirkpatrick2017a} is one of the most used regularization strategies. It is importance-based, in the sense that it computes importance coefficients for each parameter at each step and penalizes drastic changes for important parameters. 
On the other hand, Learning without Forgetting (LwF) \parencite{Li2016a} leverages a distillation loss to keep the network's output close to its previous value. \\
\textbf{Architectural strategies} try to mitigate forgetting by enhancing the model's plasticity. Typically, they expand the network by adding more units \parencite{Draelos2017a, marsland2002a}, an entirely new module \parencite{Rusu2016a, Cossu2020a}, or by expanding and then compressing the resulting architecture \parencite{Hung2019a, Srivastava2019a}. These approaches require careful management of resources to avoid high computational costs.\\
\textbf{Replay strategies} mix input patterns from the current step with patterns from previously encountered steps \parencite{Isele2018a, rolnick2019a}. Replay memory management is crucial because it is not feasible to store all the patterns from previous steps. Generative replay, instead, overcomes this problem by training a generative model (with fixed space occupancy) that provides on-demand previous patterns \parencite{Shin2017a, ven2020a, Wang2019a}.

\subsection{Deep Learning for Graphs}
\label{subsec:graph-representation-learning}

When it comes to learning from input samples represented as a graph, classical recurrent or recursive approaches cannot deal with the mutual dependencies between the nodes, as these create cycles in the structure. There is a long and consolidated history of works that discuss these problems, with some of them dating back more than twenty years ago \parencite{sperduti_supervised_1997,frasconi_general_1998,micheli_neural_2009,scarselli_graph_2009}. Nowadays, the models that can process a broad spectrum of graphs by means of local and iterative processing of information are called Deep Graph Networks\footnote{This term disambiguates the more common \quotes{Graph Neural Networks} (GNN), which refers to the work of \parencite{scarselli_graph_2009}.} (DGNs) \parencite{bacciu_gentle_2020}. Generally speaking, DGNs propagate nodes' information across the graph by stacking several graph convolutional layers on top of each other. Each layer works by aggregating each node's neighbouring information, and it ultimately produces node representations that can be used to make predictions about nodes, links, or entire graphs. For the sake of brevity, we refer the reader to recent works that summarize the state of the art \parencite{bronstein_geometric_2017,battaglia_relational_2018,bacciu_gentle_2020,wu_comprehensive_2020}.

In what follows, we describe the CL strategies and deep graph networks used to evaluate catastrophic forgetting in the domain of graph-structured data; to the best of our knowledge, this is one of the first studies to investigate this particular aspect. To keep the discussion clear, we will focus on regularization and replay strategies applied to simple architectures for graphs, deferring more complex techniques to future studies.
\section{Techniques}
\label{sec:techniques}
We now describe in more detail the CL techniques that we tested on Deep Graph Networks.

\subsection{Elastic Weight Consolidation}
Elastic Weight Consolidation \parencite{Kirkpatrick2017a} is a regularization technique which prevents changes in parameters that are important for previous steps. Formally, EWC adds a squared penalty term $\mathcal{R}$ to the classification loss at training time:
\begin{equation} \label{eq:ewc}
    \mathcal{R}(\mathbf{\Theta}, \mathbf{\Omega}) = \lambda \sum_{i=1}^{n-1} \mathbf{\Omega_i} \|\mathbf{\Theta_i} - \mathbf{\Theta_n}\|_2^2,
\end{equation}
where $\mathbf{\Theta_n}$ is the vector of parameters of current step $n$, $\mathbf{\Theta_i}$ is the vector of parameters from previous step $i$ and $\mathbf{\Omega_i}$ is the vector of parameter importances for step $i$. The hyperparameter $\lambda$ controls the trade-off between classification accuracy on current step and stability of parameters. The importance for step $n$ is computed at the end of training on step $n$, through a diagonal approximation of the Fisher Information Matrix:
\begin{equation}
\mathbf{\Omega_n}= \mathbb{E}_{(\mathbf{x}, \mathbf{y}) \in \mathcal{D}} \big[(\nabla_{\mathbf{\Theta_n}} \log p_\mathbf{\Theta_n}(\mathbf{y}|\mathbf{x}))^2 \big].
\end{equation}
The computation of importance values requires an additional pass over the training data $\mathcal{D}$ and the estimation of the log probabilities $\log p_{\mathbf{\Theta}}$ represented by the network outputs. Following \cite{Schwarz2018a}, we keep a single importance matrix for all steps, by summing the importance on the current step with the previous values. In order to prevent the unbounded growth of importance values we normalize between $0$ and $1$ when computing importance on the current step.

\subsection{Learning without Forgetting}
Learning without Forgetting (LwF) \parencite{Li2016a} is a regularization technique which preserves the knowledge of previous steps by fostering stability at the activation level through knowledge distillation \cite{hinton2015distilling}. The method adds a regularization term $\mathcal{R}$ to the loss during step $n$ as follows:
\begin{equation}
    \mathcal{R}(\mathbf{\Theta_n}, \mathbf{\Theta_{n-1}}; \mathbf{x}, \mathbf{y}) = \alpha\ \text{KL}[p_\mathbf{\Theta_n}(\mathbf{y}|\mathbf{x})\ ||\ p_\mathbf{\Theta_{n-1}}(\mathbf{y}|\mathbf{x})],
\end{equation}
where $\alpha$ controls the regularization strength. The KL-divergence term prevents current activations to diverge too much from the ones of the model at previous step.

\subsection{Replay}
Replay of previous patterns during training is a very effective technique against forgetting of existing knowledge \parencite{rolnick2019a, Aljundi2019a, hayes2019a, chaudhry2019b}. We leveraged a replay memory which stores a fixed number of patterns for each class. During training on each step, the replay memory is concatenated with the training set. The resulting dataset is shuffled and used for training the model. Therefore, replay patterns are spread uniformly over the training set. 

\subsection{Na\"ive}
The Na\"ive strategy trains the model continuously without applying any CL technique. This strategy is heavily subjected to catastrophic forgetting. Therefore, it can be used as a baseline to compare the performance of more effective CL strategies, which should perform significantly better in terms of forgetting.

\subsection{Architectural Details}

We define a graph as a tuple $g = (\mathcal{V}_g,\mathcal{E}_g,\mathcal{X}_g, \mathcal{A}_g)$ where $\mathcal{V}_g$ is the set of \emph{nodes}, $\mathcal{E}_g$ is the set of oriented \emph{edges} connecting nodes, whereas $\mathcal{X}_g$ (respectively $\mathcal{A}_g$) denotes node (edge) features. The neighbourhood $\mathcal{N}_v$ of a node $v$ is the set of all nodes $u$ for which an edge $(u,v)$ directed towards $v$ exists. 


\paragraph{Structure-agnostic Baseline}
To assess whether continual learning strategies have an impact when working with graphs, we must first devise a baseline that ignores the structural information and relies only on node features. The most common baseline we find in the literature \parencite{dwivedi_benchmarking_2020,errica_fair_2020} is a multi-layer perceptron (MLP) that is invariant to the ordering of the nodes. Formally, the baseline compute a node representation $\mathbf{h}_v$ as follows
\begin{align}
& \mathbf{h}_v = \psi(\mathbf{x}_v), \ \ x_v \in \mathcal{X}_g, \\ 
& \psi(x_v) = \mathbf{W}_L^T(\sigma(\dots(\sigma(\mathbf{W}_1^Tx_v + \mathbf{b}_1)\dots) + \mathbf{b}_L),
\label{eq:baseline-node}
\end{align}
where $\psi(\cdot)$ is an MLP of $L$ layers, the symbol $\mathbf{W}$ denotes a weight matrix and $\mathbf{b}$ is the bias. As the tasks under consideration in this paper deal with graph classification, an additional \textit{readout} phase is necessary, in which we aggregate all node representations into a single graph representation $\mathbf{h}_g$:
\begin{align}
    \mathbf{h}_g = \Psi_g \Big( \{ \mathbf{h}_v \mid v \in \mathcal{V}_g \} \Big),
\label{eq:readout}
\end{align}
where $\Psi_g$ is a permutation invariant function; in this work we will use the \textit{mean} function as the baseline's readout.

\paragraph{Deep Graph Networks}
While DGNs usually adopt the same readout scheme as the one of Equation \ref{eq:readout}, the fundamental difference lies in its graph convolutional layer. If we assume a deep network of $L$ layers, the node representation at layer ${\ell}<L$, that is, $\mathbf{h}_v^{\ell}$ is obtained by aggregating the neighbouring information of all nodes using another permutation invariant function $\Psi_n$:
\begin{align}
\mathbf{h}_v^{\ell+1} = \phi^{\ell+1} \Big(\mathbf{h}_v^\ell,\ \Psi_n(\{\psi^{\ell+1}(\mathbf{h}_u^{\ell}) \mid u \in \mathcal{N}_v\} ) \Big),
\label{eq:aggregation}
\end{align}
where $\phi$ and $\psi$ are usually implemented as linear layers or MLPs. 

In our experiments, we 
define $\Psi_n$ as the \textit{mean} operator for digit classification tasks and sum for the chemical ones.

\paragraph{Structure-preserving Regularization Loss}
We believe it is worth investigating whether a structure-preserving regularization loss such as the one of \parencite{kipf_semi-supervised_2017} affects catastrophic forgetting when used alongside the various CL strategies. The catch is that regularization will help preserve the output of previously seen classes when similar structural patterns appear in the new training samples. In general, the interplay between GRL and CL regularization strategies opens appealing research directions for the future. In case the chosen regularization does not help, this may indicate that the distribution of neighbour states of patterns belonging to a new class is radically different from those seen before.
\section{Experiments}
\label{sec:experiments}
This section provides a thorough description of the experimental details necessary to reproduce our experiments. The code is made publicly available to reproduce the results and carry out novel robust evaluations of different continual learning strategies\footnote{\url{https://github.com/diningphil/continual_learning_for_graphs}}. 
\subsection{Datasets}
\label{subsec:datasets}

The evaluation is carried out on three different large graph classification datasets. The former two, namely MNIST and CIFAR10, are the standard digit classification benchmarks used in the CL literature. However, here the digits are represented as graphs of varying dimension and shape \parencite{dwivedi_benchmarking_2020}. The nodes are \quotes{superpixels} obtained through a specific coarsening process, and the adjacency information is constructed using the $k$-nearest neighbour algorithm. We defer the specifics of this process to the original paper. The third dataset is OGBG-PPA \parencite{hu_open_2020}, a dataset of undirected protein association neighbourhoods taken from protein-protein interaction graphs. Here the task is to classify each input as one of 37 different taxonomy groups. Here, node features are missing but edges contain information. As such, we treat edges as nodes in the structure-agnostic baseline. We use the same data splits as those provided in the original papers, thus performing standard hold-out model selection and assessment. We also use the readily available version of all datasets provided by the Pytorch Geometric library \parencite{fey_fast_2019}. Table \ref{tab:dataset} summarizes some useful dataset statistics.

\begin{table}[t]
\begin{tabular}{lccc}
\toprule
                      & MNIST                         & CIFAR10                       & OGBG-PPA                    \\ \midrule
Size                  & 70000                         & 60000                         & 158100                         \\
Node Attrs.       & 3                             & 5                             & 0                              \\
Edge Attrs.       & 0                             & 0                             & 7                              \\
Classes               & 10                            & 10                            & 37                             \\
Avg $|\mathcal{V}_g|$             & 70,57                         & 117,63                        & 243,4                          \\
Avg $|\mathcal{E}_g|$             & 564,63                        & 941,07                        & 2266,1                         \\
Data Split  & 55K/5K/15K                    & 45K/5K/15K                    & 49\%/29\%/22\%                 \\
Class Split & \multicolumn{1}{l}{2+2+2+2+2} & \multicolumn{1}{l}{2+2+2+2+2} & \multicolumn{1}{l}{17+5+5+5+5} \\ \bottomrule
\end{tabular}
\caption{Summary of the datasets' statistics. \quotes{Class split} refers to how we group classes in the Split CL experiment.}
\label{tab:dataset}
\end{table}

\begin{figure*}[t]
    \centering
        \begin{subfigure}[t]{0.33\textwidth}
        \centering
        \includegraphics[width=\textwidth]{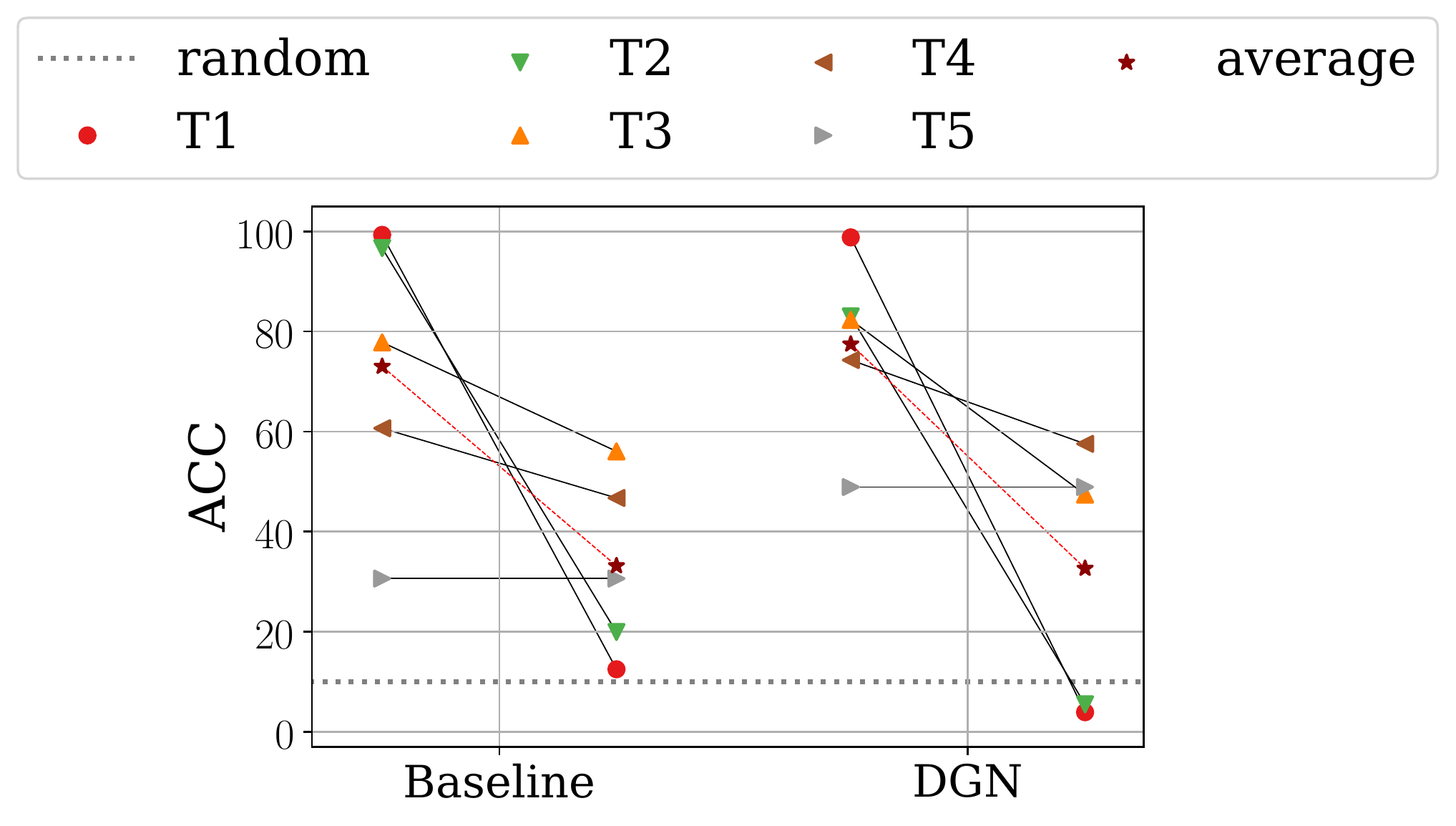}
        \caption{MNIST + LWF}
        \end{subfigure}
        \begin{subfigure}[t]{0.33\textwidth}
        \centering
        \includegraphics[width=\textwidth]{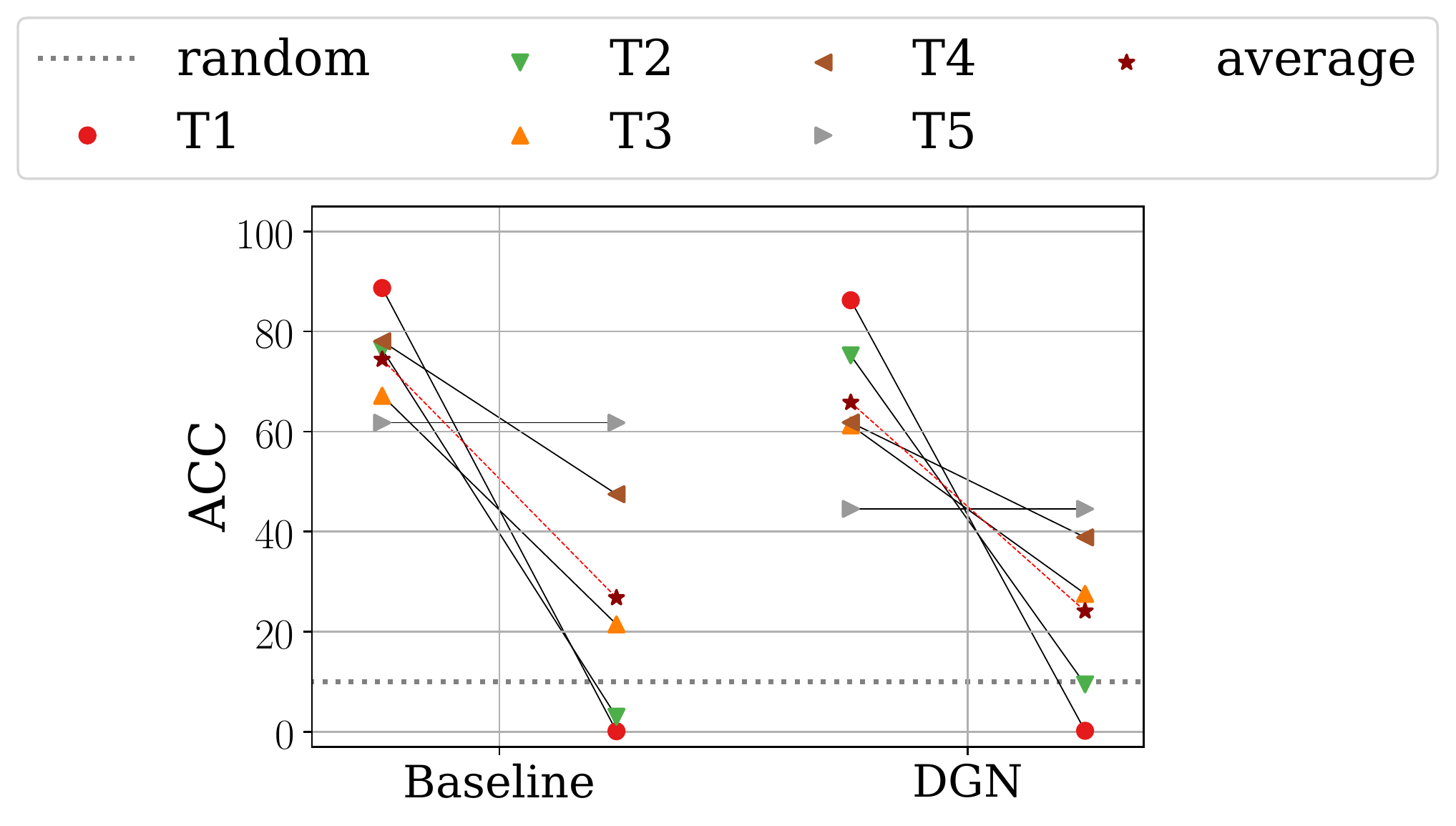}
        \caption{CIFAR10 + LWF}
        \end{subfigure}
        \begin{subfigure}[t]{0.33\textwidth}
        \centering
        \includegraphics[width=\textwidth]{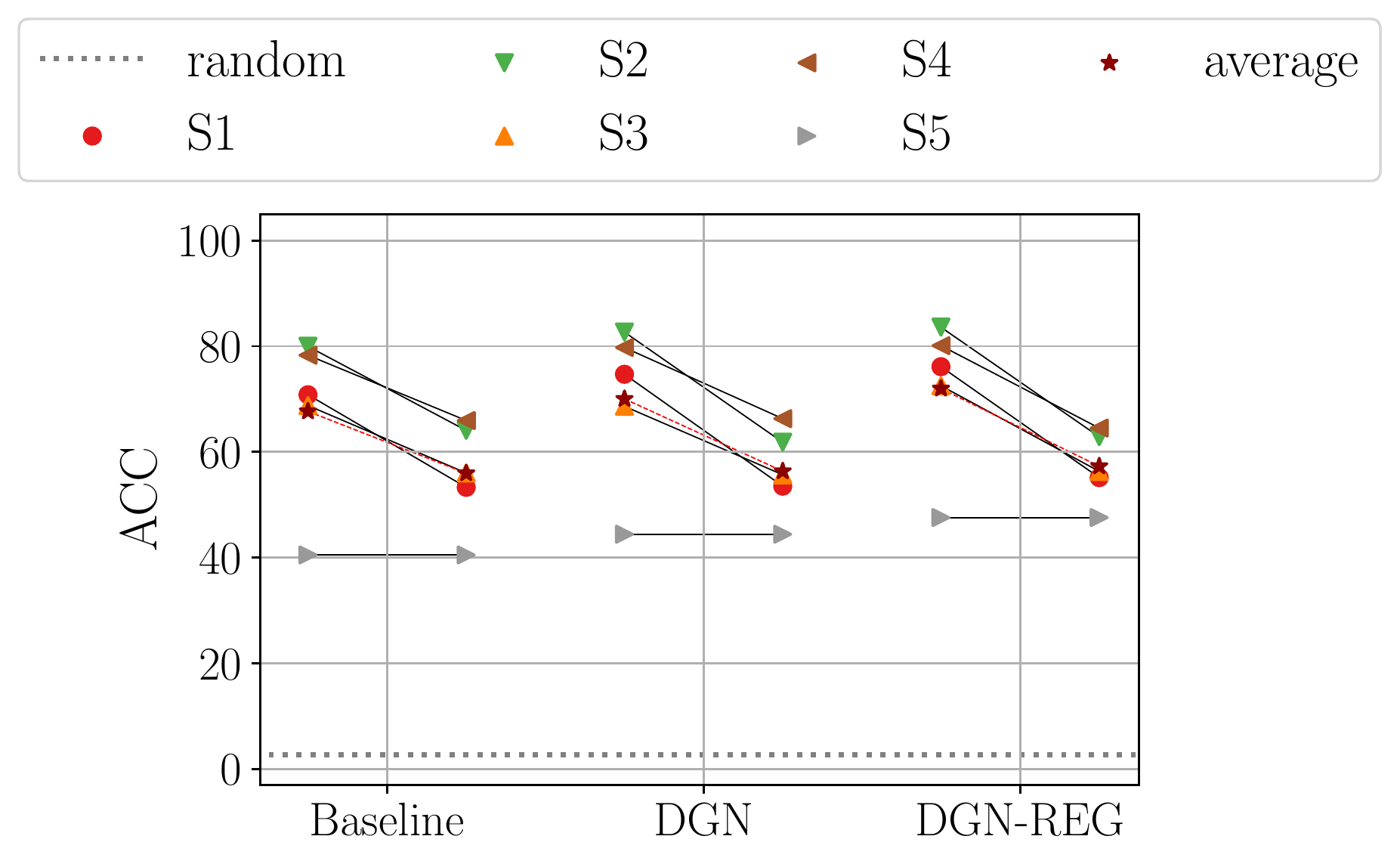}
        \caption{OGBG-PPA + REPLAY}
        \end{subfigure}
    \caption{Paired plots showing the ACC on each step for different models. Each column refers to a model and it is composed by pairs of connected points. Each pair refers to a specific step. The leftmost point in the pair represents ACC after training on that specific step. The rightmost point represents ACC after training on all steps. The more vertical the line connecting the points, the larger the forgetting effect. The dashed horizontal line indicates the performance of a random classifier. The red star represents the average performance over all steps.}
    \label{fig:paired}
\end{figure*}
\subsection{Experimental Setup}
\label{subsec:setup}
We evaluate each model in the class-incremental scenario, a popular continual learning setting where new classes arrive over time. When a new steps arrives, the model is trained on the new data without using data from the previous steps (except for the replay buffer). We use single-head models, where the entire output layer is used at each step. Table \ref{tab:dataset} shows the class splits for each dataset. To select the best hyperparameters for each strategy (see Appendix A), we perform the model selection on a separate validation set. The best hyperparameters found during the model selection are used for the model assessment on the test set. We monitor the metric $\text{ACC} =\frac{1}{T} \sum_{t=1}^T R_{T,t}$, introduced in \cite{Lopez-Paz2017a}, where $R_{T,t}$ is the accuracy on step $t$ after training on step $T$. \\
We report the average ACC and its standard deviation computed over $5$ runs. We evaluate the performance by computing the mean accuracy over all the steps after training on all steps.
\section{Results}
\label{sec:results}
The empirical results suggest that Deep Graph Networks trained continuously are subjected to catastrophic forgetting of previous knowledge. Table \ref{tab:results} reports the average ACC across all steps. We also extend the results presented in \cite{Lesort2020b} to Deep Graph Networks: importance-based regularization strategies are not able to prevent forgetting in class-incremental scenarios. In fact, in our experiments EWC always performs comparably to the \textit{Na\"ive} strategy.\\
Interestingly, Deep graph networks do not provide significant performance improvements with respect to a structure-agnostic baseline. This is a surprising result, which might have two complementary explanations. The first is that the neighboring states' distribution of different classes varies, thus making the previously trained graph convolutions inadequate for subsequent tasks. The second, instead, relates to the nature of the class-incremental scenario. Since the model sees few classes at a time, each training task becomes so simple that the model ends up relying on node features only to discern between the two classes. This is confirmed by the fact that, when encouraged to retain structural information via the regularization term, DGN shows a slight increase in performance with the replay strategy. We believe that addressing both points in more detail could constitute interesting future work at the intersection of the two research fields.

Not all regularization strategies are, however, subjected to forgetting. In fact, we show that LwF is able to recover part of the original knowledge, outperforming both Na\"ive and EWC. 
We also found LwF to be very sensitive to the choice of the hyperparameters (Appendix \ref{app:lwf}). In particular, the softmax temperature and the hyperparameter $\alpha$, which controls the amount of knowledge distillation heavily influence the final performance. 
This limits the applicability of LwF in real world applications due to the constraints of model selection in continual learning scenarios \cite{Chaudhry2019a}.

Replay strategy is considered among the strongest CL strategies available. In our experiments, replay consistently outperforms all the other strategies. Appendix \ref{app:replay} expands on replay by showing the final performance under different replay memory sizes. Deep graph networks and baseline models require a comparable amount of replay to obtain the same level of performance. Therefore, replay seems to behave as a good model-agnostic strategy even in the domain of graphs. 

\begin{table}[t]
\small
\centering
{\renewcommand{\arraystretch}{1.5} 
\begin{tabular}{llcllll}
\toprule & \multirow{2}{*}{\textbf{Model}} & \multicolumn{4}{c}{\textbf{Strategy}}                                                                                             \\
                &                      & \multicolumn{1}{c}{\textbf{Na\"ive}} & \multicolumn{1}{c}{\textbf{EWC}} & \multicolumn{1}{c}{\textbf{Replay}} & \multicolumn{1}{c}{\textbf{LWF}} \\ \midrule
\multirow{3}{*}{\rotatebox[origin=c]{90}{\small\textsc{MNIST}}}
& Baseline        & \meanstd{19.56}{0.1} & \meanstd{19.39}{0.1} & \meanstd{86.13}{4.5} & \meanstd{33.16}{13.1} &  \\ 
& DGN             & \meanstd{19.19}{0.1} & \meanstd{18.95}{0.3} & \meanstd{79.52}{1.9} & \meanstd{32.64}{5.0} &  \\ 
& DGN+reg         & \meanstd{19.31}{0.1} &  \multicolumn{1}{c}{---} & \meanstd{81.42}{2.4} & \multicolumn{1}{c}{---} &  \\ \midrule
\multirow{3}{*}{\rotatebox[origin=c]{90}{\small\textsc{CIFAR10}}}
& Baseline        & \meanstd{17.49}{0.1} & \meanstd{17.49}{0.1} & \meanstd{42.87}{3.7} & \meanstd{26.77}{5.1} &  \\
& DGN             & \meanstd{17.11}{0.2} & \meanstd{17.10}{0.2} & \meanstd{39.55}{2.3} & \meanstd{24.13}{4.1} &  \\
& DGN+reg         & \meanstd{17.13}{0.1} &  \multicolumn{1}{c}{---} & \meanstd{46.61}{3.5} & \multicolumn{1}{c}{---} &  \\ \midrule
\multirow{3}{*}{\rotatebox[origin=c]{90}{\small\textsc{OGBG-PPA}}}
& Baseline        & \meanstd{14.53}{0.5}   & \meanstd{13.90}{0.8}  & \meanstd{55.96}{3.0}    & \meanstd{20.83}{6.1}  &                                  \\
& DGN             & \meanstd{14.47}{0.3} & \meanstd{14.15}{0.5} & \meanstd{56.34}{2.5} & \meanstd{18.46}{5.4} &                                  \\
& DGN+reg         & \meanstd{15.18}{0.8} &  \multicolumn{1}{c}{---} & \meanstd{57.27}{3.2}    &  \multicolumn{1}{c}{---}       &                                 \\ \bottomrule
\end{tabular}
}
\caption{Mean accuracy and mean standard deviation (in parenthesis) among all steps. Replay results are related to memory size of $1000$. Results are averaged over $5$ final runs. We treat the regularization loss as a separate strategy.}
\label{tab:results}
\end{table}
\section{Conclusions}
\label{sec:conclusions}
Learning from a data stream in a continual fashion is fundamental for real-world applications. In this paper, we show that deep graph networks suffer from catastrophic forgetting in class-incremental settings. Interestingly, while graph networks outperform feedforward baselines during offline training, our results show that this advantage disappears in continual learning scenarios. While our preliminary results suggest that regularization techniques for DGN may help, the results are still far from the performance achieved in the offline setting. This suggests that more research is needed to explore whether alternative DGN or regularization techniques may be better able to exploit the graph structure and learn robust features.  We release our code and baseline models hoping to foster additional research in this direction.

\bibliographystyle{ACM-Reference-Format}
\bibliography{cl_grl.bib}

\begin{figure*}[t]
    \centering
        \begin{subfigure}[t]{0.33\textwidth}
        \centering
        \includegraphics[width=\textwidth]{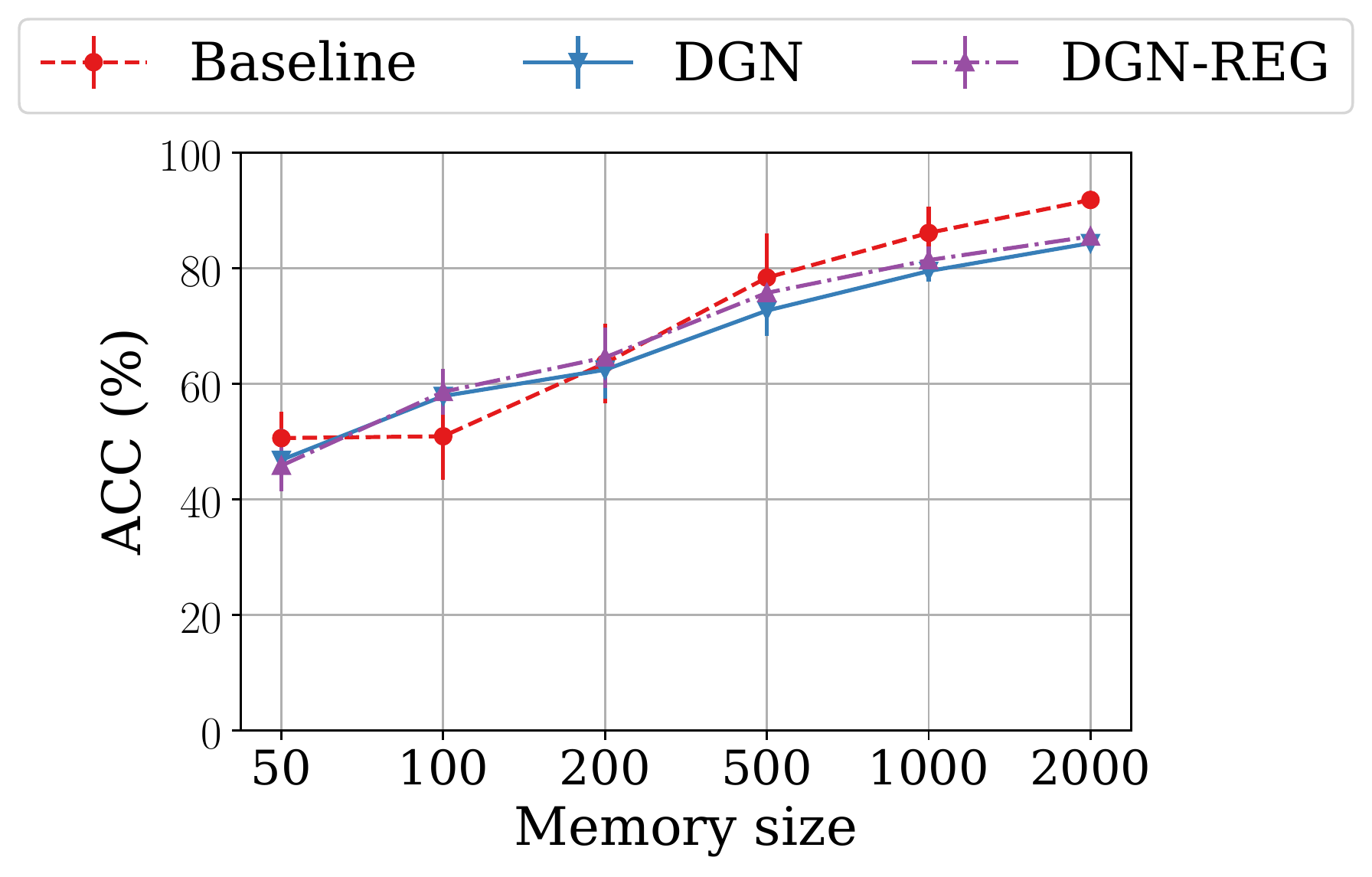}
        \caption{MNIST}
        \end{subfigure}
        \begin{subfigure}[t]{0.33\textwidth}
        \centering
        \includegraphics[width=\textwidth]{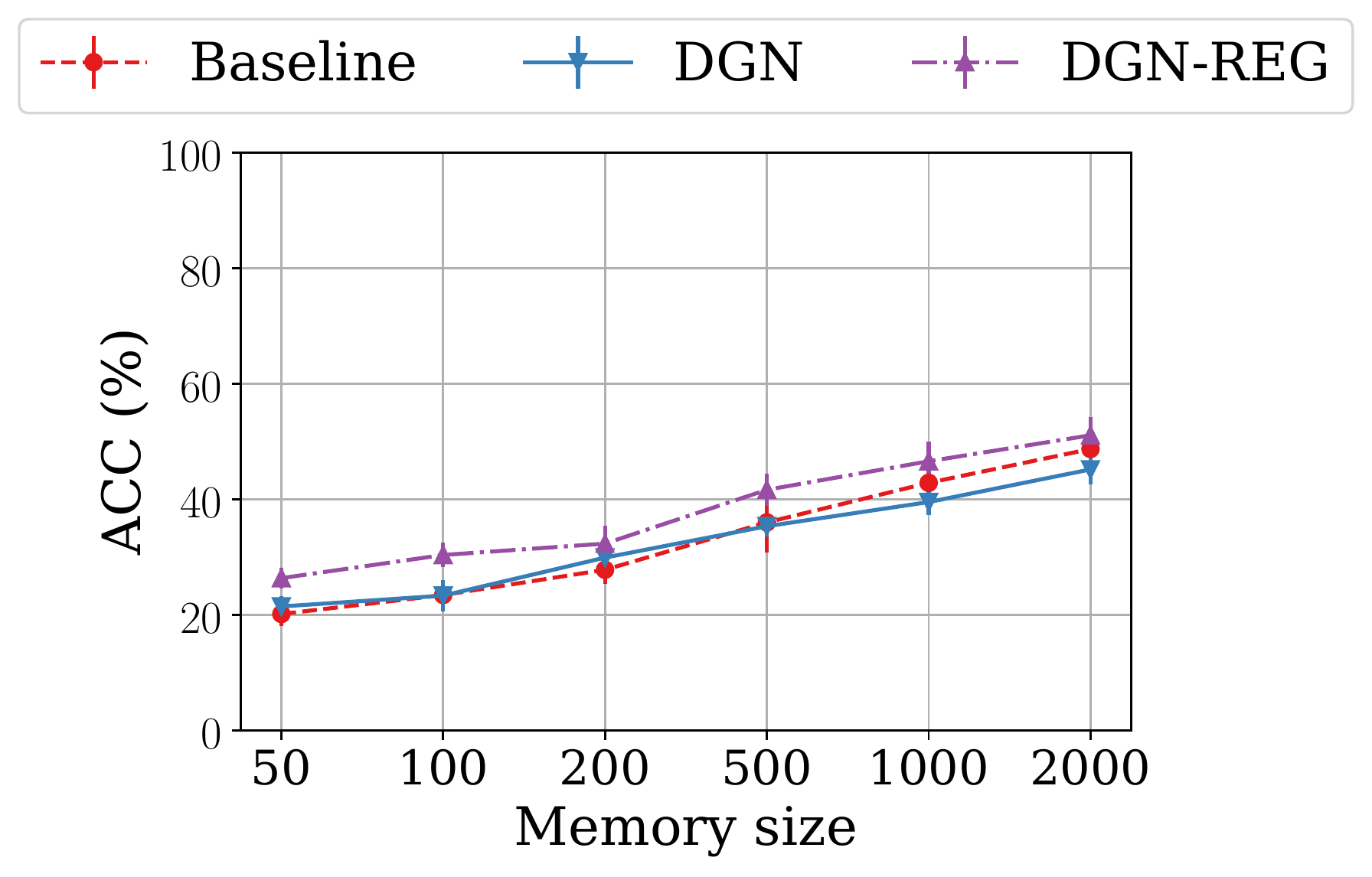}
        \caption{CIFAR10}
        \end{subfigure}
        \begin{subfigure}[t]{0.33\textwidth}
        \centering
        \includegraphics[width=\textwidth]{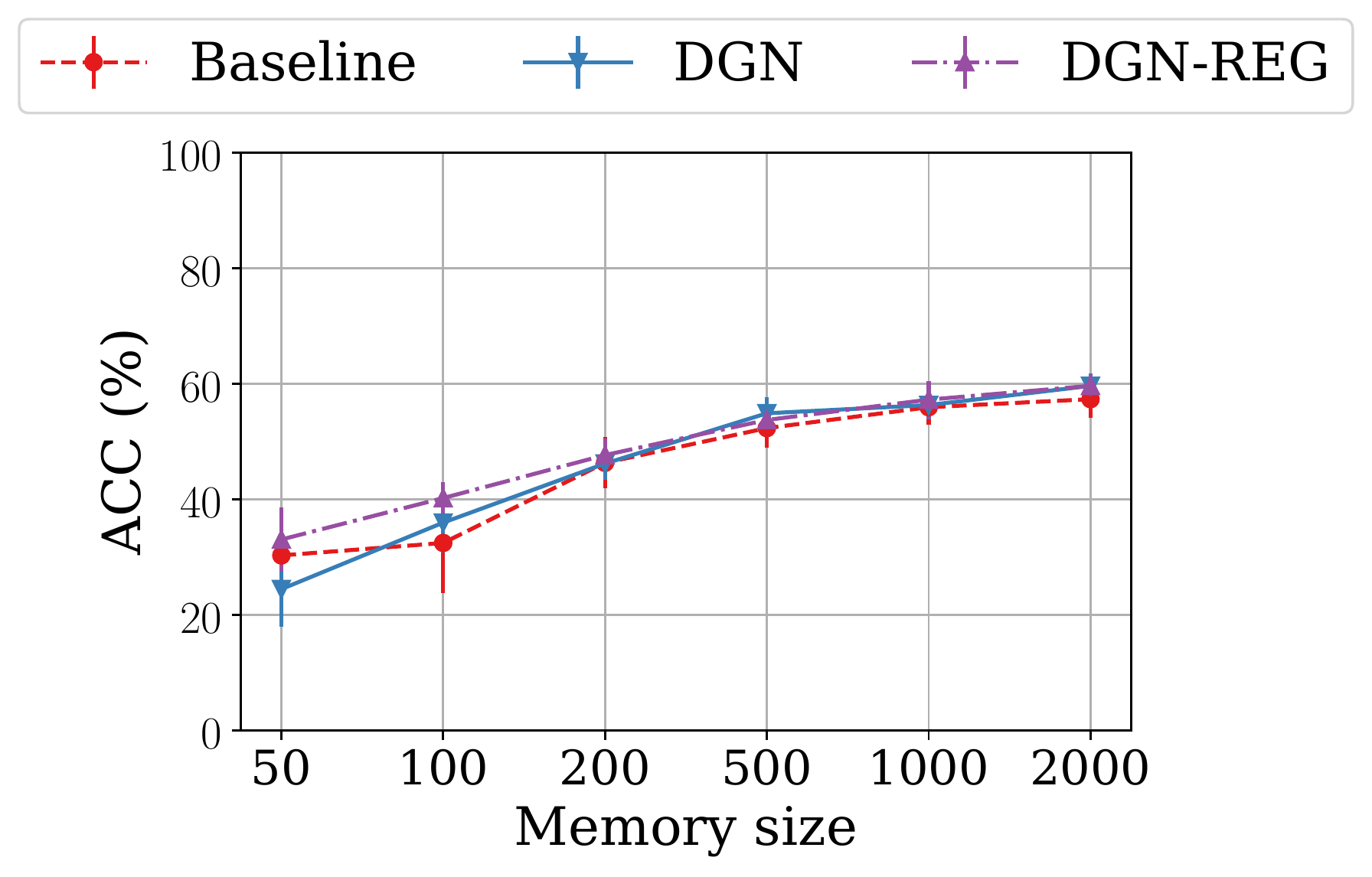}
        \caption{OGBG-PPA}
        \end{subfigure}
    \caption{ACC for increasing replay memory size.}
    \label{fig:replay}
\end{figure*}

\appendix
\section{Hyper-parameters}
\label{subsec:baselines-hyperparameters}
We perform model selection on the validation set using a grid-search strategy for all the implemented models. Regardless of the dataset or continual learning technique used, we selected the number of layers in $\{2,4\}$ for the DGN and 4 for the baseline. In both cases, the dimension of the hidden layer was chosen in  $\{64,128\}$. The number of epochs was set to 200 (patience = 20) for the Baseline and to 1000 for DGN and DGN+Reg (patience = 50). The learning rate was set to 0.001, and the optimizer chosen was Adam. We used the "sum" version of the EWC combined with normalized importance scores. Being LWF very sensible to the hyper-parameters, we chose $\alpha \in \{0.5, 1., 2.\}$ and the temperature in $\{0.5, 1., 2.\}$.

\section{Additional replay experiments} \label{app:replay}
Figure \ref{fig:replay} shows ACC values for increasing replay memory sizes.

\section{Sensitivity of LwF to hyperparameters} \label{app:lwf}
\begin{figure}[t]
    \centering
    \includegraphics[width=0.3\textwidth]{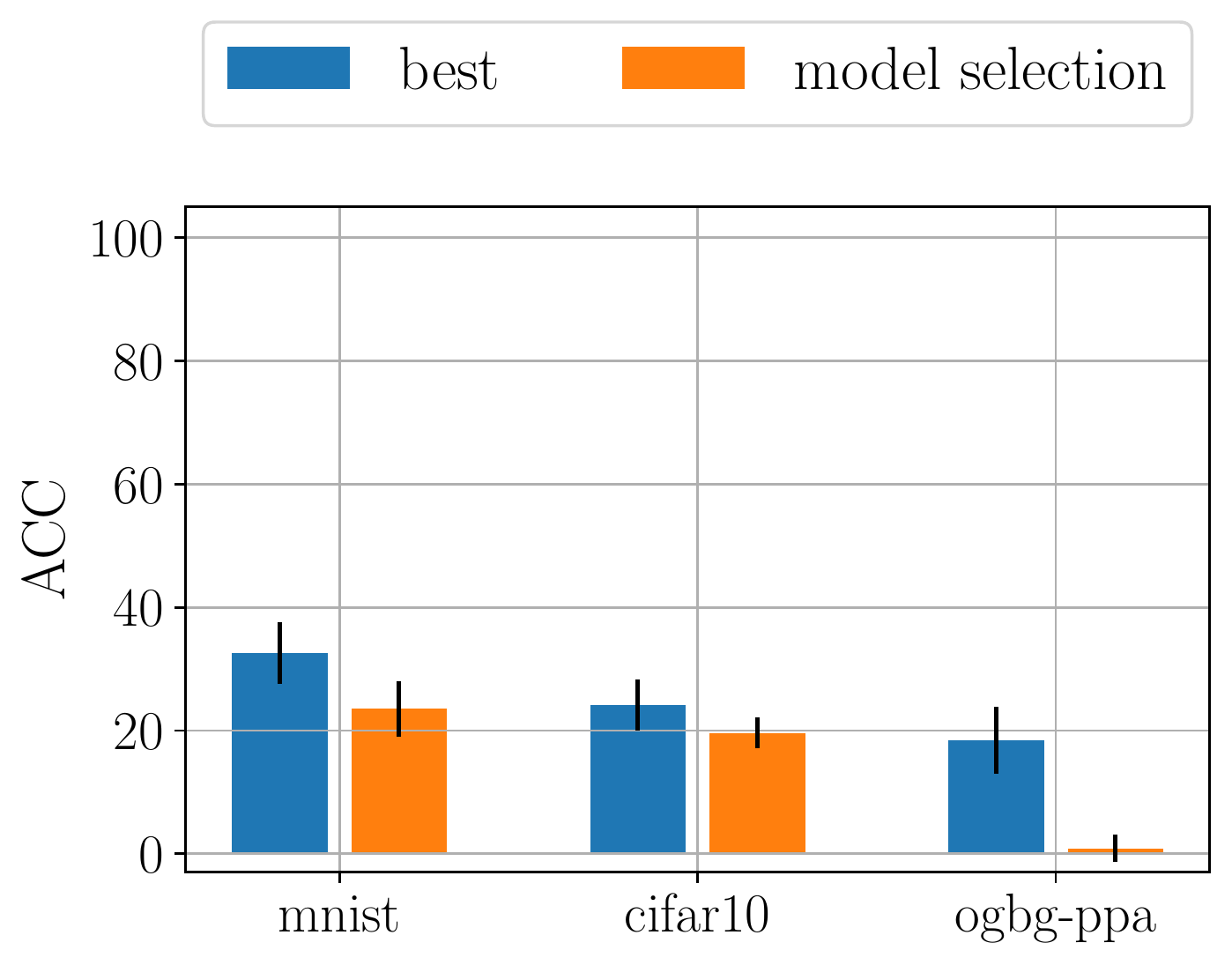}
    \caption{Comparison of performances between model selection (averaged across all configurations) and model assessment (averaged across 5 final training runs). The difference highlights the sensitivity of LwF to the choice of hyperparameters.}
    \label{fig:lwf}
\end{figure}

We briefly show the sensitivity of LwF to the choice of hyperparameters (Figure \ref{fig:lwf}). We compute the mean ACC and its standard deviation across all runs of model selection. Then, we compare the results with the best performance we found during model assessment. The difference highlights the sensitivity of Lwf to the hyper-parameters. 

\section{Visualization of results with paired plots}
\begin{figure*}[t]
    \centering
        \begin{subfigure}[t]{0.33\textwidth}
        \centering
        \includegraphics[width=\textwidth]{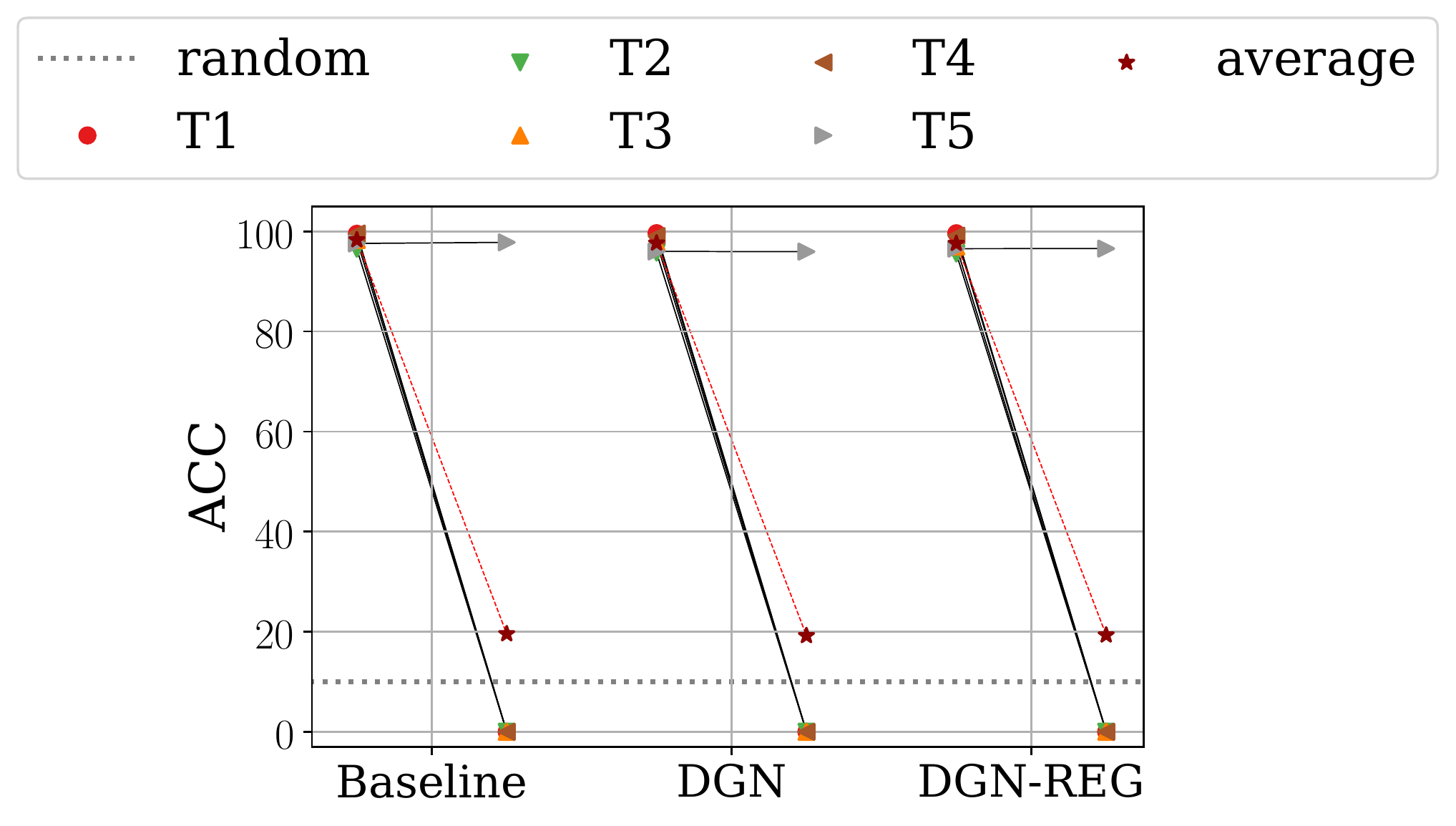}
        \caption{MNIST + Na\"ive}
        \end{subfigure}
        \begin{subfigure}[t]{0.33\textwidth}
        \centering
        \includegraphics[width=\textwidth]{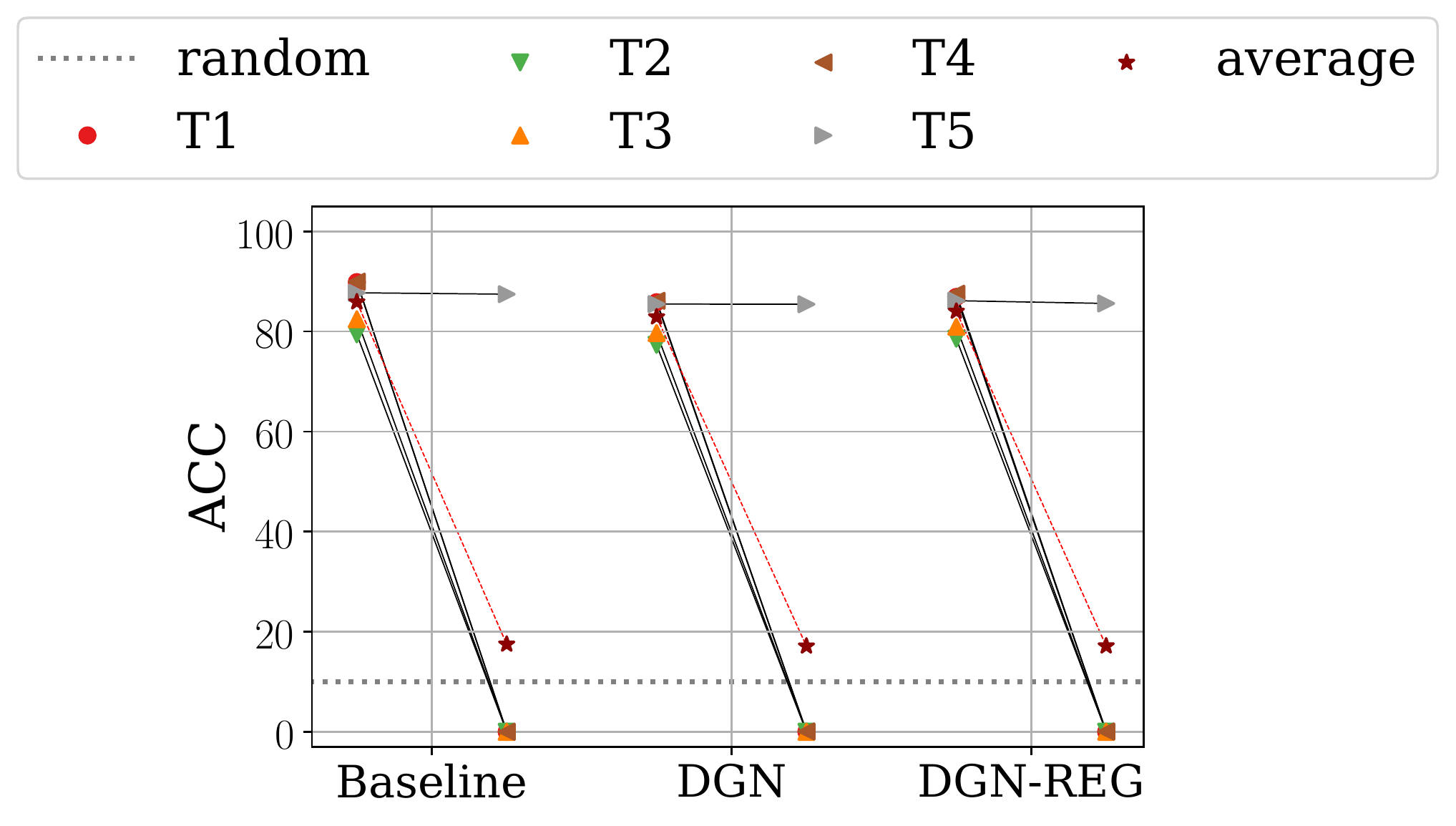}
        \caption{CIFAR10 + Na\"ive}
        \end{subfigure}
        \begin{subfigure}[t]{0.33\textwidth}
        \centering
        \includegraphics[width=\textwidth]{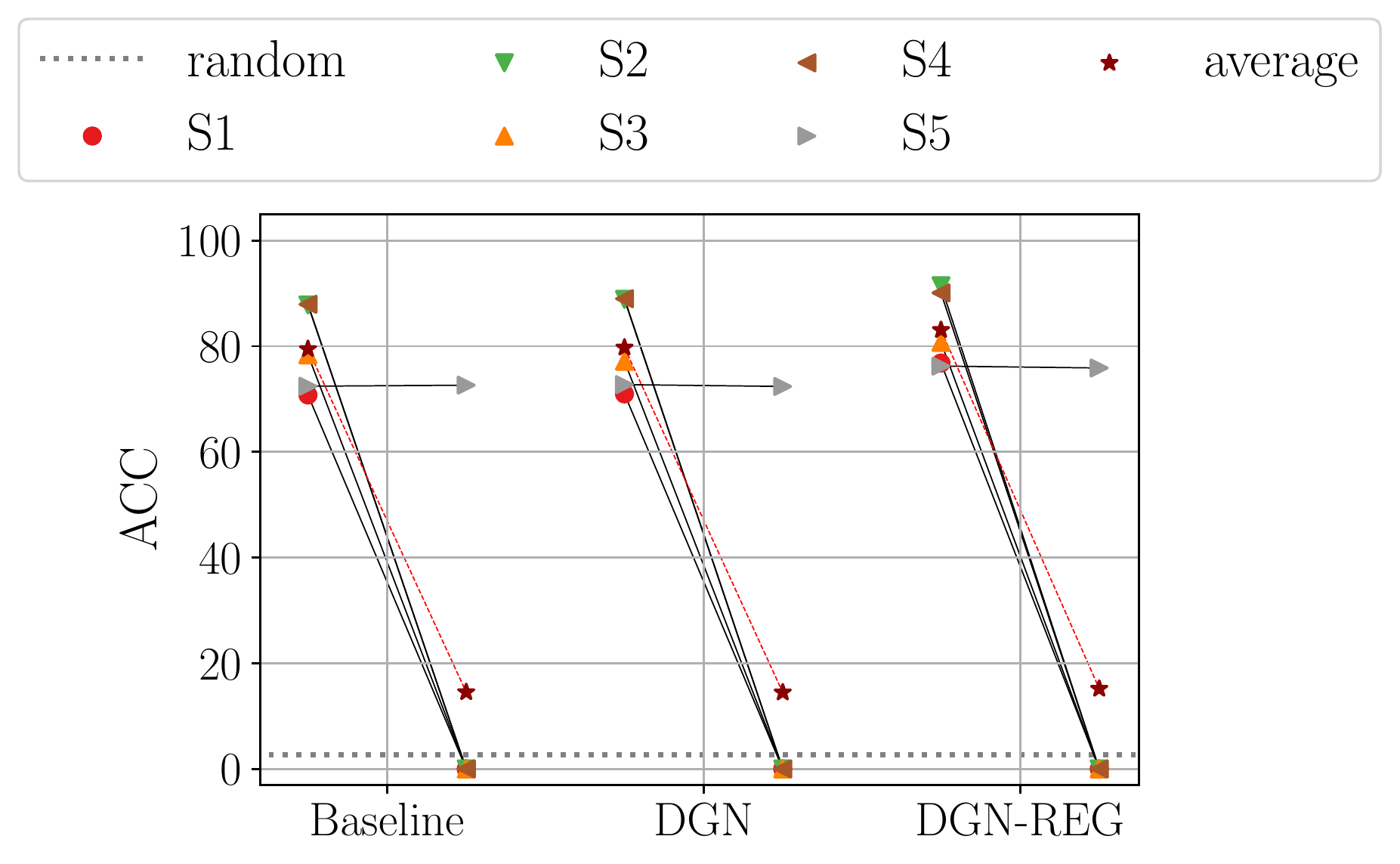}
        \caption{OGBG-PPA + Na\"ive}
        \end{subfigure}
        \begin{subfigure}[t]{0.33\textwidth}
        \centering
        \includegraphics[width=\textwidth]{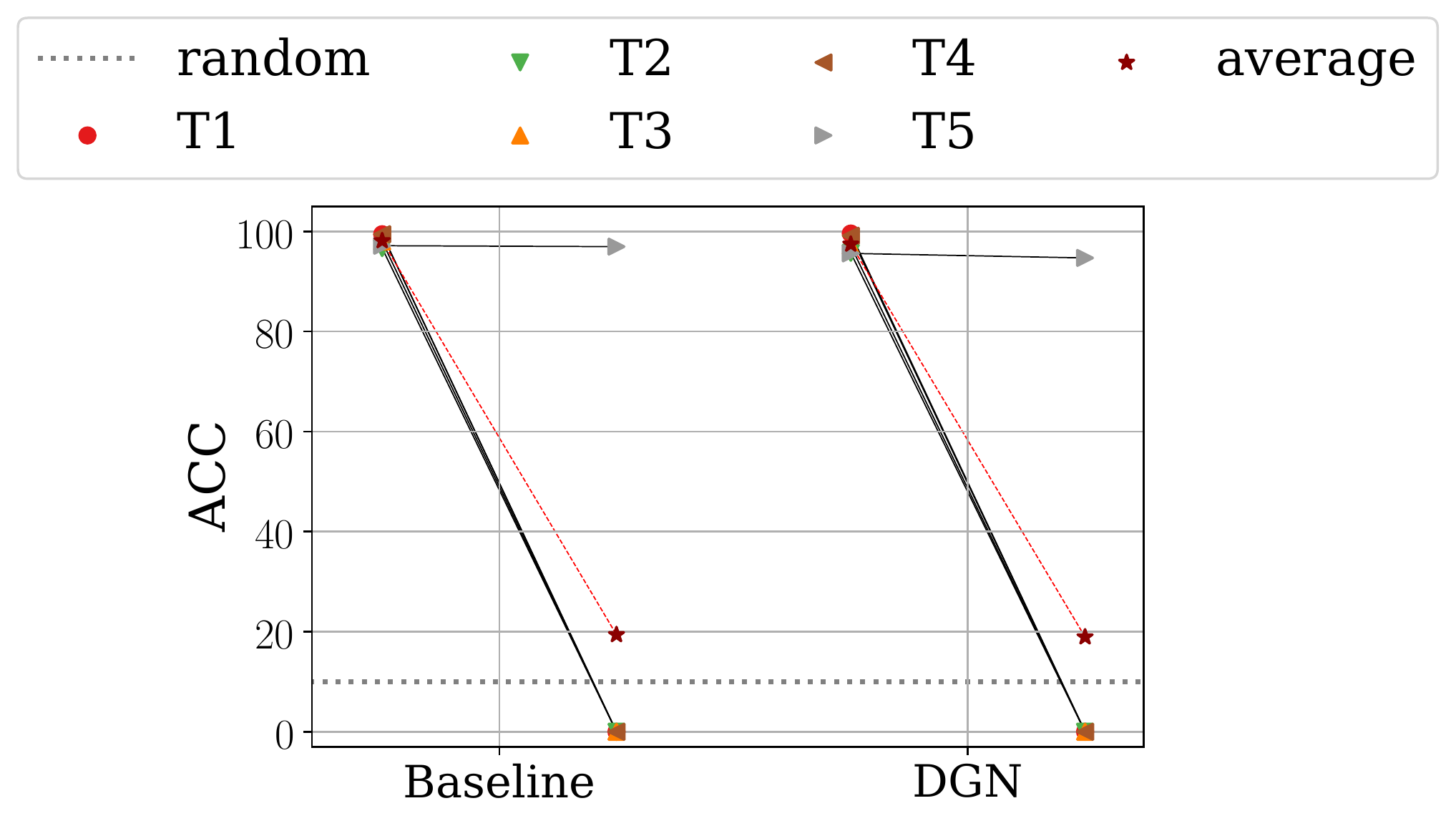}
        \caption{MNIST + EWC}
        \end{subfigure}
        \begin{subfigure}[t]{0.33\textwidth}
        \centering
        \includegraphics[width=\textwidth]{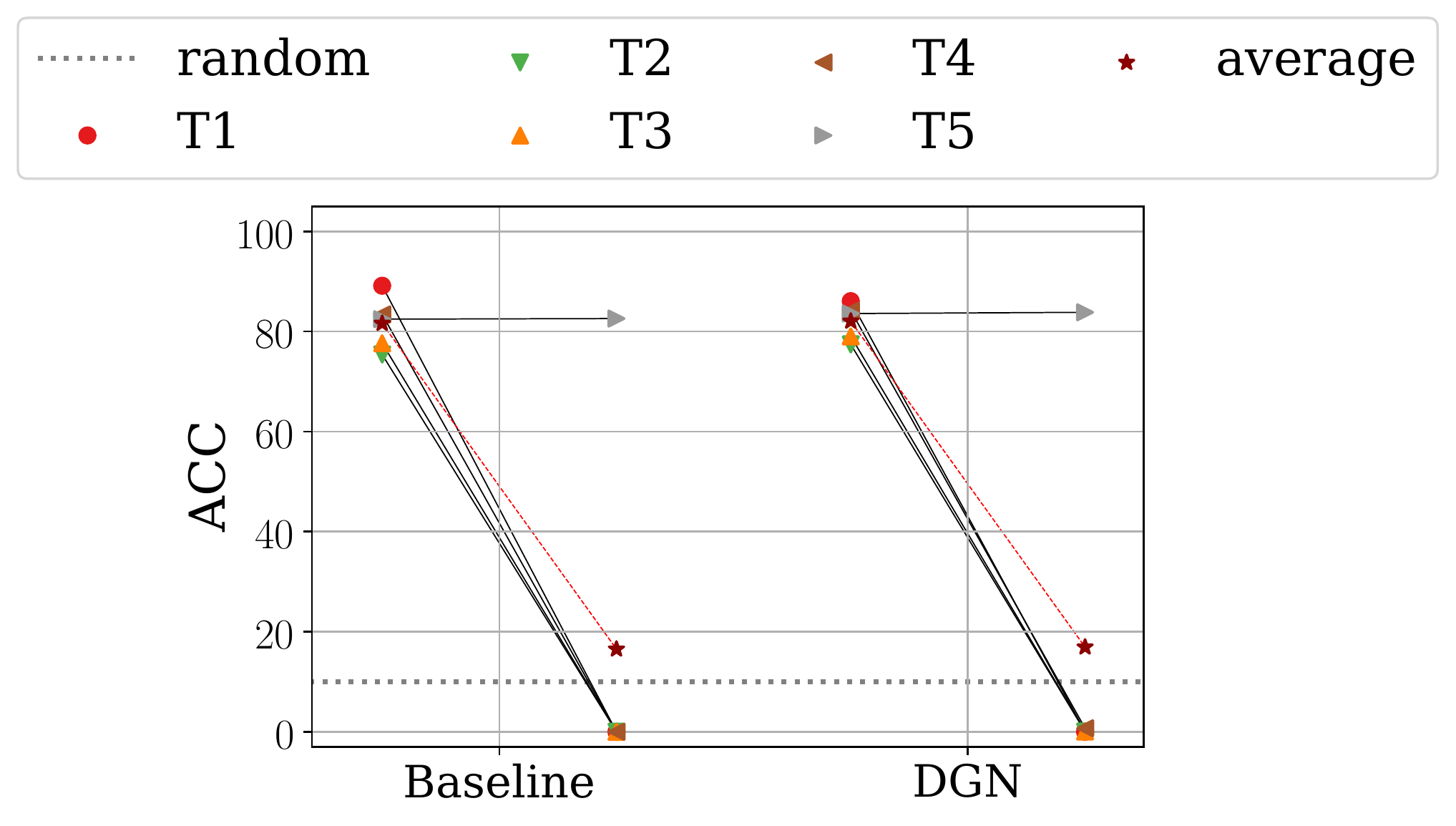}
        \caption{CIFAR10 + EWC}
        \end{subfigure}
        \begin{subfigure}[t]{0.33\textwidth}
        \centering
        \includegraphics[width=\textwidth]{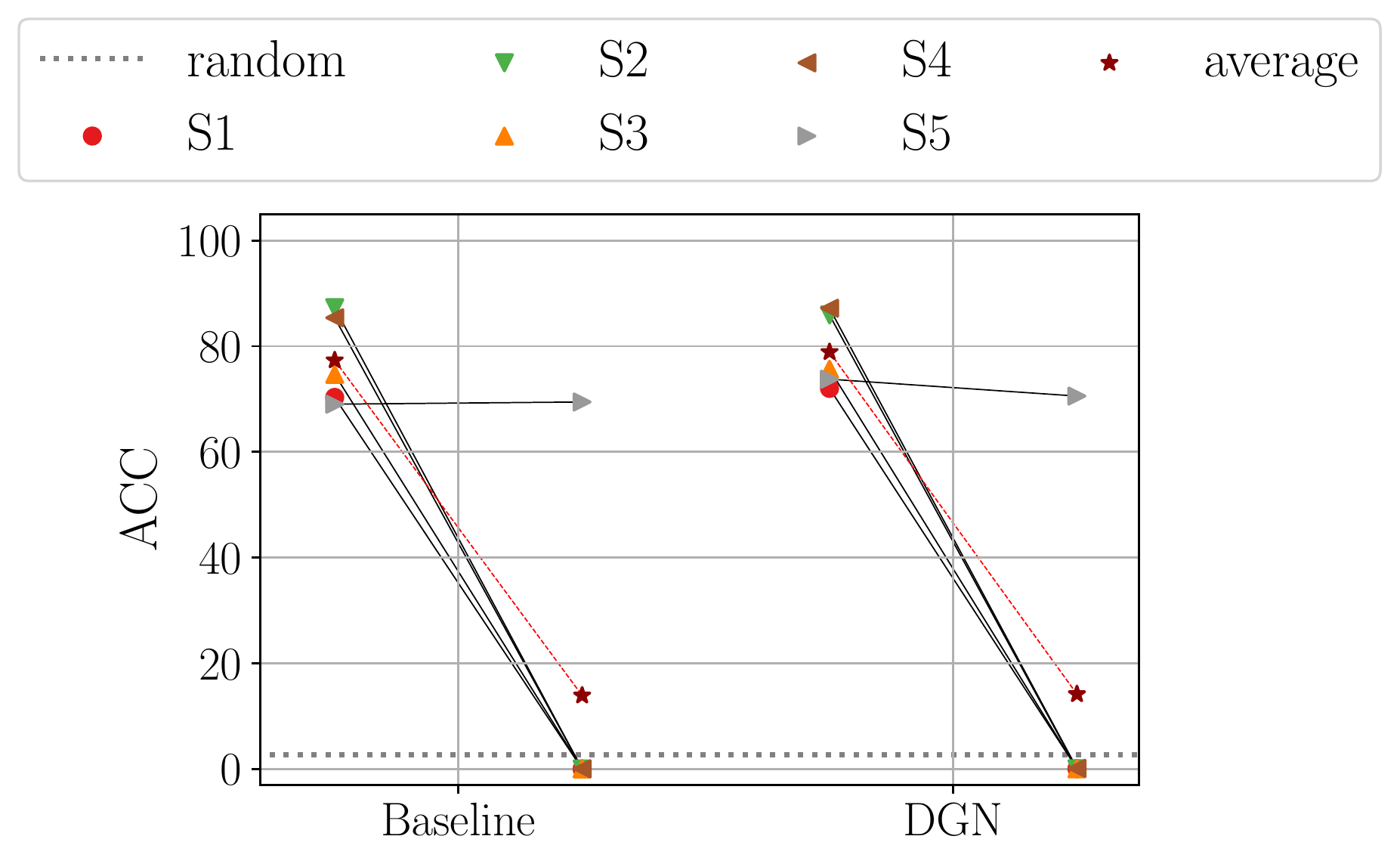}
        \caption{OGBG-PPA + EWC}
        \end{subfigure}
        \begin{subfigure}[t]{0.33\textwidth}
        \centering
        \includegraphics[width=\textwidth]{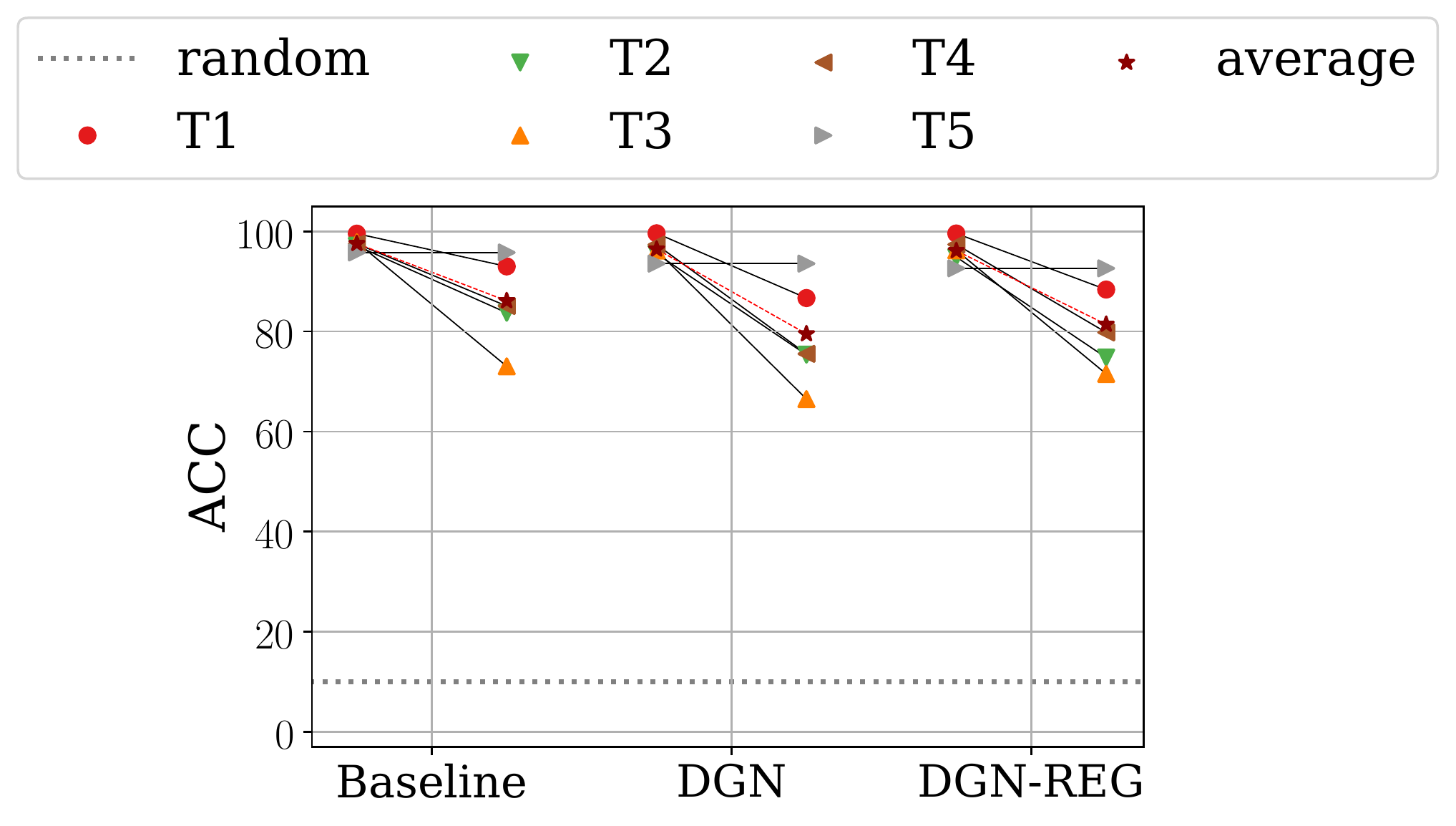}
        \caption{MNIST + REPLAY}
        \end{subfigure}
        \begin{subfigure}[t]{0.33\textwidth}
        \centering
        \includegraphics[width=\textwidth]{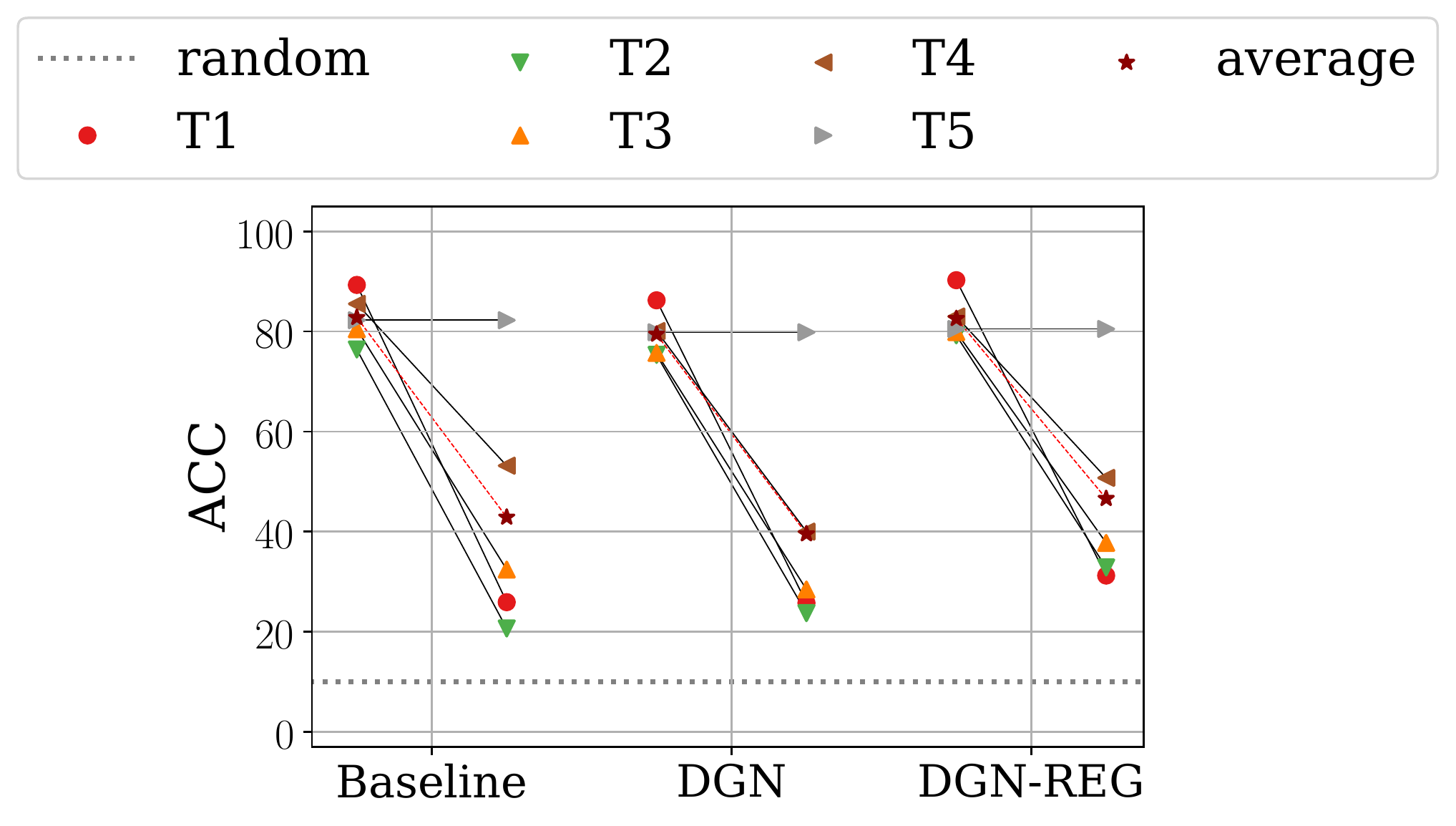}
        \caption{CIFAR10 + REPLAY}
        \end{subfigure}
        \begin{subfigure}[t]{0.33\textwidth}
        \centering
        \includegraphics[width=\textwidth]{img/pairedplots/REHEARSAL_1000_ogbg_ppa_paired_plot.pdf}
        \caption{OGBG-PPA + REPLAY}
        \end{subfigure}   
        \begin{subfigure}[t]{0.33\textwidth}
        \centering
        \includegraphics[width=\textwidth]{img/pairedplots/LWF_MNIST_paired_plot.pdf}
        \caption{MNIST + LwF}
        \end{subfigure}
        \begin{subfigure}[t]{0.33\textwidth}
        \centering
        \includegraphics[width=\textwidth]{img/pairedplots/LWF_CIFAR10_paired_plot.pdf}
        \caption{CIFAR10 + LwF}
        \end{subfigure}
        \begin{subfigure}[t]{0.33\textwidth}
        \centering
        \includegraphics[width=\textwidth]{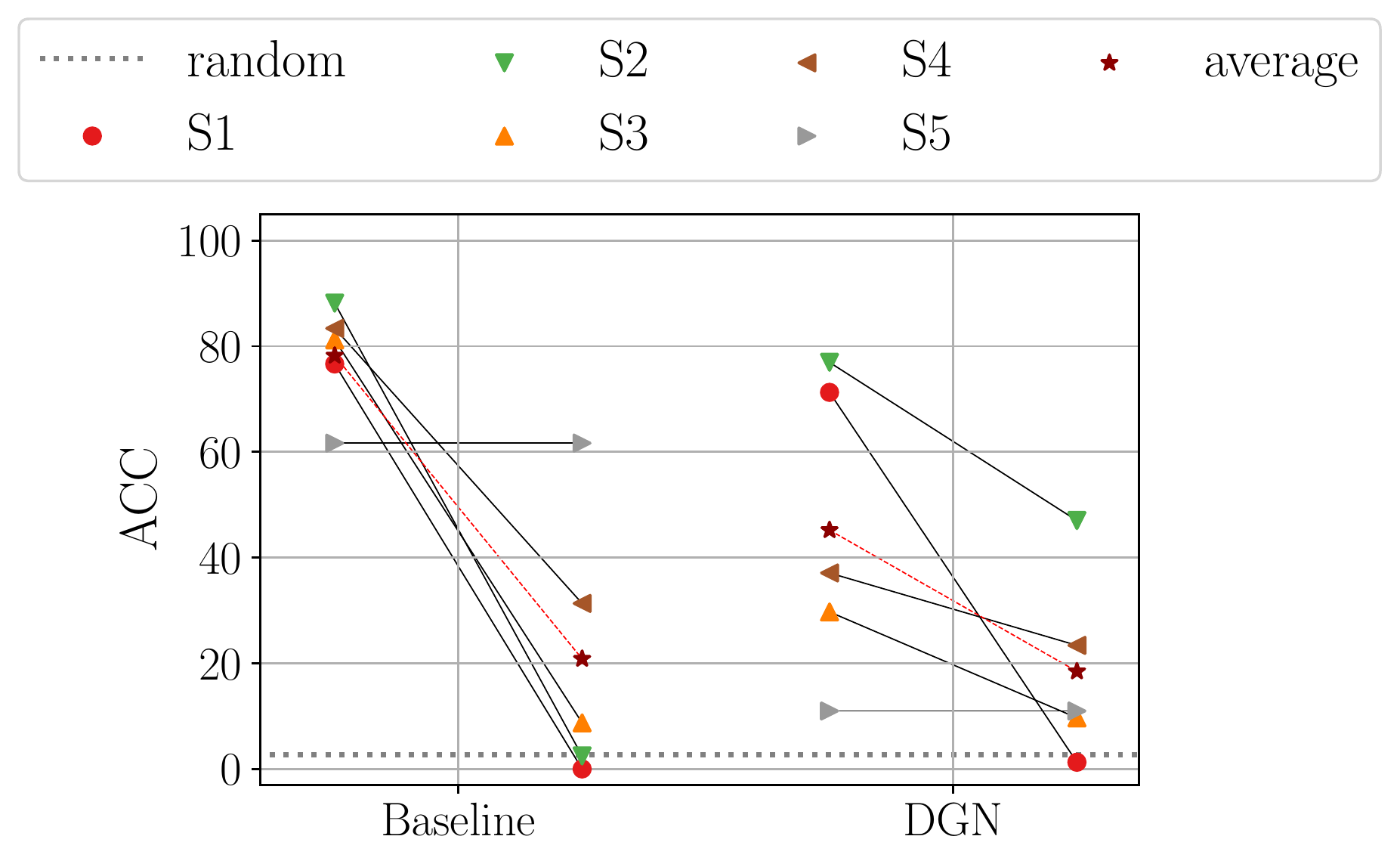}
        \caption{OGBG-PPA + LwF}
        \end{subfigure}
    \caption{Additional paired plots for the experiments. Refer to Figure \ref{fig:paired} for a description of paired plots.}
    \label{fig:additionalpaired}
\end{figure*}

For completeness, Figure \ref{fig:additionalpaired} reports the paired plots for all the CL techniques and datasets tested in this work.

\end{document}